\documentclass[letterpaper, 10 pt, conference]{ieeeconf}  % Comment this line out if you need a4paper

\IEEEoverridecommandlockouts                              % This command is only needed if 
\usepackage{graphics} % for pdf, bitmapped graphics files
\usepackage{epsfig} % for postscript graphics files
\usepackage{amsmath} % assumes amsmath package installed
\usepackage{amssymb}  % assumes amsmath package installed
\usepackage{bm} % added by Kentaro UNO 2020/12/08 to use the \bm font

\usepackage[flushmargin,hang]{footmisc} % added by Kentaro UNO 2020/11/05 to make the letter of \thanks to aligned to left
\usepackage{flushend} % added by Kentaro UNO 
\usepackage{comment}	

\usepackage{siunitx} % to use \SI command to visualize the unit
\usepackage{pgfplots}
\pgfplotsset{compat=newest}
\usetikzlibrary{plotmarks}
\usetikzlibrary{arrows.meta}
\usepgfplotslibrary{patchplots}
\usepackage{booktabs} % For better table formatting
\usepackage{grffile}
\usepackage{url}
\pgfplotsset{plot coordinates/math parser=false}
\newlength\fwidth
\newlength\fheight

\usepackage{cite}

\title{\LARGE \bf
Enhancing Autonomous Manipulator Control with Human-in-loop for Uncertain Assembly Environments} 
\author{Ashutosh Mishra$^{1*}$, Shreya Santra$^{1}$, Hazal Gozbasi$^{1}$, Kentaro Uno$^{1}$ and Kazuya Yoshida$^{1}$% <-this % stops a space
\thanks{
This work was supported by JST Moonshot R\&D Program, Grant Number JPMJMS223B.
    }% 
\thanks{$^{1}$A. Mishra, S. Santra, H. Gozbasi,  K. Uno, and K. Yoshida are with the Space Robotics Lab. (SRL) in Department of Aerospace Engineering, Graduate School of Engineering, Tohoku University, Sendai 980--8579, Japan. (E-mail: \tt{ashutosh.mishra@dc.tohoku.ac.jp})  }% 
\thanks{
\textit{$^{*}$The corresponding author is Ashutosh Mishra}
    }%
}%

\begin{document}

\maketitle
\thispagestyle{empty}
\pagestyle{empty}

%%%%%%%%%%%%%%%%%%%%%%%%%%%%%%%%%%%%%%%%%%%%%%%%%%%%%%%%%%%%%%%%%%%%%%%%%%%%%%%%

\begin{abstract}
%%%%%%%%%%%%%%To be updated
This study presents an advanced approach to enhance robotic manipulation in uncertain and challenging environments, with a focus on autonomous operations augmented by human-in-the-loop (HITL) control for lunar missions. By integrating human decision-making with autonomous robotic functions, the research improves task reliability and efficiency for space applications. The key task addressed is the autonomous deployment of flexible solar panels using an extendable ladder-like structure and a robotic manipulator with real-time feedback for precision. The manipulator relays position and force-torque data, enabling dynamic error detection and adaptive control during deployment. To mitigate the effects of sinkage, variable payload, and low-lighting conditions, efficient motion planning strategies are employed, supplemented by human control that allows operators to intervene in ambiguous scenarios. Digital twin simulation enhances system robustness by enabling continuous feedback, iterative task refinement, and seamless integration with the deployment pipeline.
The system has been tested to validate its performance in simulated lunar conditions and ensure reliability in extreme lighting, variable terrain, changing payloads, and sensor limitations. %\todo{add a sentence on the results obtained}

\end{abstract} 

%%%%%%%%%%%%%%%%%%%%%%%%%%%%%%%%%%%%%%%%%%%%%%%%%%%%%%%%%%%%%%%%%%%%%%%%%%%%%%%%

\section{Introduction}\label{introduction}
%%%%%%%To be updated
The increasing focus on long-term lunar missions and future human settlements in recent years highlights the need for autonomous robots in infrastructure development. At the lunar poles, the predominantly horizontal sunlight necessitates the vertical arrangement of solar and communication towers \cite{lunar_tower}. Deploying these structures with robots requires precision but is complicated by extreme lighting, soil sinkage, and sensor limitations, making the environment highly unpredictable despite the repetitive nature of the tasks \cite{camille_2024}, \cite{luca_2025}. To address these challenges, Human-in-the-Loop (HITL) control enhances autonomous robotic manipulation by enabling human intervention in unforeseen scenarios, improving reliability while preserving autonomy for routine operations. Dynamic replanning then uses real-time sensor data to avoid collisions and keep the task on track.

\begin{figure}[t]
    \centering
    \includegraphics[width=0.9\linewidth]{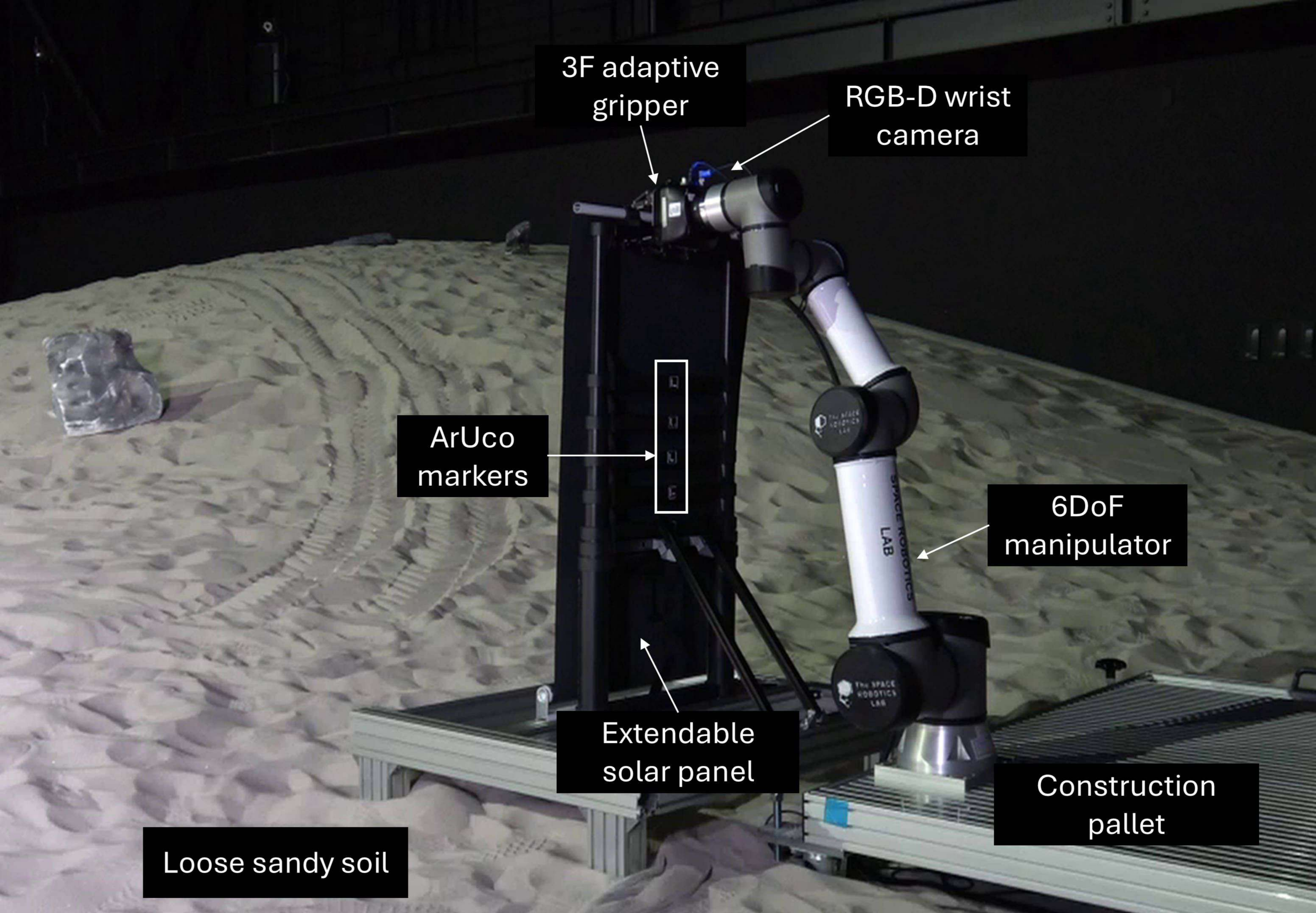}
    %\hspace{0.5cm} % Adds horizontal space between the images
    \caption{Setup for the solar panel deployment at the experiment site with low-illumination and loose sinking soil}
    \label{fig:experiment_setup}
    %vspace{-5mm}
\end{figure}

The key contributions of this study include testing a robotic manipulator in a sandy field environment, evaluating its performance with dynamic payloads, low-illumination and variable terrains. To facilitate this, an extendable tower-like structure was deployed, using vision feedback based on fiducial ArUco markers for precise positioning as shown in Fig.~\ref{fig:experiment_setup}. The system underwent extensive real-world testing at the \textit{Advanced Facility for Space Exploration (TansaX)}  \cite{Tansa_X} at the Japan Aerospace Exploration Agency (JAXA) Sagamihara Campus, to validate its functionality in artificial lunar environments.

%%%%%%%%%%%%%%%%%%%%%%%%%%%%%%%%%%%%%%%%%%%%%%%%%%%%%%%%%%%%%%%%%%%%%%%%%%%%%%%%
\section{Related Works}\label{relatedwork}
%% literature review

While fully autonomous systems are necessary for future, semi-autonomous operations are crucial for ensuring safety in uncertain environments like space. In such settings, human-machine interaction plays a vital role in preventing accidents \cite{hitl}. For example, closed-loop operator feedback systems used during space rendezvous and docking rely on sensors to provide precise guidance, assisting astronauts with proximity maneuvers and aligning the spacecraft's pose with the target aboard space stations. The study in \cite{vision_based_docking} demonstrates how human-in-the-loop teleoperation—utilizing a joystick, real-time video feedback, and trajectory prediction—enables successful docking with uncooperative, rotating targets, even under time delays. Such feedback greatly improves safety and docking reliability \cite{assisted_docking}.

%\todo{Human in the loop for autonomous construction }
In order to conduct assembly and construction tasks successfully, it is necessary to identify and localize the goal positions. Dynamic viewpoint impacts the control techniques on teleoperation performance in construction tasks, as shown in \cite{robotic_welding}. In \cite{crane_lifting}, Goh et al., propose a human-in-the-loop simulation with an immersive interface for crane lift planning and hazard identification. Accurate simulations enable iterative development by incorporating environmental factors to evaluate performance, address challenges, and refine control algorithms before actual field deployment.
Some of the most commonly used simulators in robotics community are Gazebo, Webots, CoppeliaSim, Unity3D and Isaac Sim  \cite{robotics_simulators_comparison}. Isaac Sim developed by NVIDIA Omniverse uses the PhysX engine and has high graphical fidelity often used for photorealistic simulations \cite{robotic_simulators}. In \cite{Jose2024}, Moyo et al., demonstrate a 6-DoF manipulator that autonomously adapts to dynamic part positions and orientations using AprilTag-based goal updates, enabling assembly without reprogramming.

In our work, we utilize the fiducial ArUco markers \cite{garrido2014automatic}, for precise localization and payload tracking, along with the Isaac Sim platform for simulations.

%%%%%%%%%%%%%%%%%%%%%%%%%%%%%%%%%%%%%%%%%%%%%%%%%%%%%%%%%%%%%%%%%%%%%%%%%%%%%%%%
\section{Methodology}\label{methodology}

\subsection{System Overview}\label{system_overview}
This study presents an autonomous system for deploying flexible solar panels in extraterrestrial environments, focusing on lunar infrastructure. The system consists of an extendable ladder-like structure with flexible panels and a 6-DoF manipulator equipped with an adaptive three-finger gripper, operating on a loose sandy surface. Real-time sensory feedback and HITL control enhance adaptability and reliability under variable conditions.

The manipulator’s kinematic chain enables precise trajectory planning for sequential alignment, gripping, and elevation of the structure. An RGB-D camera supplies real-time 3D pose estimation and an end-effector force-torque sensor supports  adaptive control. ArUco markers affixed to each ladder step ensure precise localization within the vision-based control system, maintaining accuracy under varying illumination conditions. The HITL functionality addresses environmental uncertainties that surpass autonomous control thresholds. Real-time position and force data reach the operator interface for manual overrides during critical stages. This interactive feedback loop reduces error propagation and bolsters mission success rates. Vision-based proximity detection and collision-avoidance algorithms further safeguard both the manipulator and the solar panel array.

\subsection{Hardware Configuration}\label{hardware}
The autonomous system employs the UR16e 6-DoF industrial robotic arm of \SI{16}{kg} payload capacity, \SI{900}{mm} reach, and ±\SI{0.05}{mm} repeatability \cite{ur16e_specs}, mounted on a fixed aluminium frame. High-torque servos ensure precise joint motion, while the manipulator’s modular design facilitates integration with diverse end-effectors and sensors \cite{ur16e_specs}. An in-built six-axis force-torque sensor enables adaptive grip control, dynamic payload adjustment, and real-time anomaly detection.

Attached at the end-effector is the Robotiq 3F adaptive gripper, which supports a grip payload of \SI{10}{kg} and accommodates varying geometries \cite{Robotiq-3F}. Three articulated fingers 
%with adjustable grip forces 
ensure secure handling of ladder steps without structural damage. The gripper’s compliance further compensates for misalignments or irregularities, enhancing manipulation precision.

For vision-based sensing, two Intel RealSense D435i RGB-D cameras \cite{Realsense_D435i} are used, one as a wrist-eye camera to detect ArUco markers and the other as a third-eye view camera, providing accurate localization, pose estimation, and proximity detection for obstacles like ladders, terrain, and partly deployed solar panels. %Its compact size and low power consumption make it ideal for space-constrained robotics. 

All hardware components are integrated using a centralized control architecture with real-time data communication, ensuring coordinated and reliable autonomous deployment.

\subsection{Software and Control Framework} \label{software}

\subsubsection{Vision-feedback and Pose Estimation}
The vision-feedback and pose estimation system uses real-time RGB-D images for robust localization. ArUco markers on each step provide visual cues for estimating step pose via homography and perspective transformations\cite{zhang2023homography}.
%\cite{perspective_transformation}
The transformation from the marker to the camera coordinate frame is given by:
\begin{equation}
T_{\rm cam}^{\rm marker} = K \begin{bmatrix} R & t \\ 0 & 1 \end{bmatrix}
\end{equation}
where $K$ is the camera intrinsic matrix, $R$ is the rotation matrix, and $t$ is the translation vector. The overall pose of each ladder step in the world frame is then derived by:
\begin{equation}
T_{\rm world}^{\rm step} = T_{\rm world}^{\rm cam} \cdot T_{\rm cam}^{\rm marker}
\end{equation}

Here, $T_{world}^{step}$ and $T_{world}^{cam}$ are the homogeneous transformation matrices representing the pose of a ladder step and camera in the world coordinate frame respectively. $T_{cam}^{marker}$ is the transformation from the ArUco marker to the camera frame, estimated via vision. A \textbf{Coarse-to-Fine Adaptive Thresholding} mechanism enhances fiducial marker detection through two stages.In the \textbf{broad tuning} stage, random thresholds $c$ are applied across frames until detection succeeds, allowing adaptation to unknown lighting. The subsequent \textbf{fine-tuning} stage samples candidate thresholds near the successful $c$ and evaluates them over $N$ frames. Let $D_i(c) \in \{0, 1\}$ indicate detection success in frame $i$ with threshold $c$. The optimal threshold is:
\begin{equation}
c_{\text{optimal}} = \arg\max_c \sum_{i=1}^{N} D_i(c)
\end{equation}
This two-stage strategy improves robustness by adapting to illumination variations using exploratory sampling followed by local optimization.

\subsubsection{Motion Planning and Collision Avoidance}
The motion planning framework ensures precise manipulation of ladder steps. The UR16e robotic arm’s joint configurations are determined via inverse kinematics:
\begin{equation}
   \boldsymbol{\theta}= f^{-1}(T_d)
\end{equation}
where \boldsymbol{$\theta$} are the joint angles and $T_d$ is the desired end-effector transformation. Configurations are verified to avoid singularities and respect mechanical limits, ensuring stable deployment.
%Motion planning computes a collision-free path from the arm’s home pose to each ladder-step grasp pose. 
 %Although the UR16e has no non-holonomic limits, it must respect joint limits, workspace bounds, and soil sinkage. 

The Rapidly-exploring Random Trees (RRT)  \cite{RRTs} planner efficiently explores the configuration space by random sampling:
\begin{equation}
    T = RRT(q_{\rm init}, q_{\rm goal})
\end{equation}
where $q_{\rm init}$ and $q_{\rm goal}$ are the initial and goal configurations. Basic RRT was chosen for its low computational overhead, which suits our real-time requirements in structured environments with minimal static obstacles.
Real-time feedback from vision and force-torque sensors refines the robot's path, improving collision avoidance and adaptability to the lunar surface. %This along with HITL corrections refine the path to keep manipulation safe and accurate on unstructured ground.
To mitigate trajectory deviations caused by soil sinkage, a single-axis Sequential Motion Planning (SMP) strategy is implemented, which is a deliberate motion decomposition technique designed to reduce multi-axis coupling effects that cause instability on deformable terrain. The manipulator executes discrete motions sequentially along the $y$-axis (lateral direction, left-right relative to the robot base) for alignment, followed by the $x$-axis (forward direction, toward the ladder) for secure gripping, and finally the $z$-axis (vertical direction, upward from the ground) for lifting. This axis-specific planning approach minimizes diagonal force components, thereby reducing tilting on deformable terrain and improving positional accuracy and deployment fidelity.

\subsubsection{Dynamic Payload Updates and Human-in-the-Loop}
As each ladder step is lifted, the manipulator's payload parameters are updated in real time to reflect the changing lifting mass:
\begin{equation}
%m_{\text{dynamic}} = m_{\text{initial}} + \Delta m
m_{\rm dynamic} = m_{\rm initial} + \Delta m
\end{equation}
where, $m_{dynamic}$: Current payload after lifting, $m_{initial}$: Payload before the current step, $\Delta m$: Mass of the newly lifted ladder step.

The measured joint and external torques enable recalculation of the effective center of mass ($\rm COM$) and inertia ($I$):
\begin{equation}
COM = \frac{\sum_{i=1}^{n} m_i x_i}{\sum_{i=1}^{n} m_i}
\end{equation}
where, $COM$: Combined center of mass, $m_i$: Mass of the $i$-th component, $x_i$: Position of the $i$-th component.

After updating these parameters, the robotic arm aligns its end-effector with the ladder step's tilt. The gripper secures the step, while a force-torque sensor monitors forces and payload dynamics. Sudden force changes trigger anomaly detection for stability. The 3F gripper’s internal grip force is not directly measured for this task, $F_g$ is adjusted via wrist sensor data:
\begin{equation}
F_g = F_{\text{desired}} + k_f \bigl(F_{\text{measured}} - F_{\text{desired}}\bigr),
\end{equation}
protecting the structure from excessive grip force. Here, $F_g$: Final grip force applied, $F_{desired}$: Desired grip force, $F_{measured}$: Force sensed by the wrist sensor, $k_f$: Gain factor for force correction.

In HITL mode, the operator confirms the self-locking mechanism after each lift. A successful confirmation advances the process, while failure triggers automatic pose adjustments. This feedback loop enhances deployment precision, with an audible click indicating secure locking.

\subsection{Simulated Digital Twin and Continuous Integration}\label{simulations}

\begin{figure}[t]
    \centering
    \includegraphics[width=\linewidth]{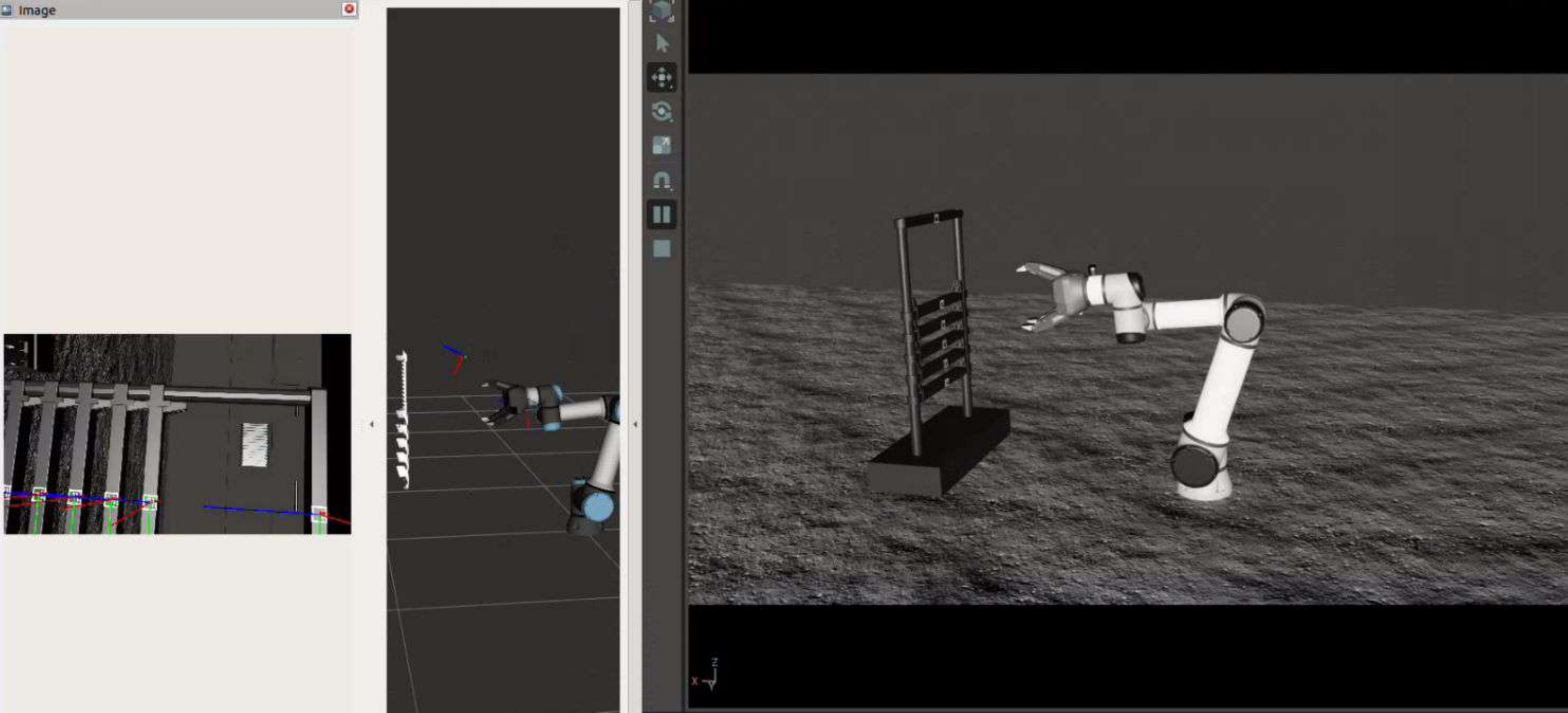}
    \caption{Rviz visualization and Digital twin simulations in IsaacSim}
    \label{fig:digital-twin}
\end{figure}

To establish the virtual simulation platform, the LunaLab setup available in OmniLRS \cite{JUN} based on NVIDIA IsaacSim was utilized, which closely replicates the sandy experimental area at JAXA. The digital twin as shown in Fig.~\ref{fig:digital-twin}, serves as a real-time operational model for precise, collision-free manipulation, optimizing path planning for efficiency and safety. Configured with real-world parameters and dynamic payload characteristics, it enables algorithm validation without hardware risk.

Before field testing, the algorithm's robustness was evaluated through simulations of varying structure positions relative to the robot. Five trials were conducted for each arm-to-ladder distance, recording elapsed time and number of deployed steps. With the manipulator at $(0,0,0)$ and the ladder initially at $(0.87, 0.05, 0.05)$, positions were varied along the $x$ and $y$ axes, as shown in Table \ref{sim_table}.  These configurations were later tested at the analogue site to determine optimal ladder placement.
%\todo{add Hazal's success table}. 

\begin{table}[h]
\caption{Simulation to assess success rates at various arm-to-ladder distances.}
\label{sim_table}
    \centering
    \begin{tabular}{|c|c|c|c|c|} % 5 columns
        \toprule
        \textbf{x (m)} & \textbf{y (m)} & \textbf{z (m)}  & \textbf{Avg. time}  & \textbf{Avg. number of}\\
        & & & \textbf{elapsed (s)} & \textbf{steps deployed} \\
       \midrule
        x = 0.87 & y = 0.05 & z = 0.05 & 1.24 & All 5 \\\hline
        x + 0.05 & y & z & 1.32 & All 5 \\\hline
        x - 0.05 & y & z & 1.32 & All 5 \\ \hline
        x + 0.05 & y + 0.02 & z & 1.28 & All 5 \\\hline
        x + 0.05 & y - 0.02 & z & 1.32 & All 5 \\ \hline
        x - 0.05 & y + 0.02 & z & 1.38 & All 5 \\ \hline
        x - 0.05 & y - 0.02 & z & 1.45 & Only 4 \\\hline
        x & y + 0.02 & z & 1.27 & All 5 \\ \hline
        x & y + 0.02 & z & 1.24 & All 5 \\ 
        \bottomrule
    \end{tabular}
\end{table}

\begin{figure}[b]
    \centering
    \includegraphics[width=0.8\linewidth]{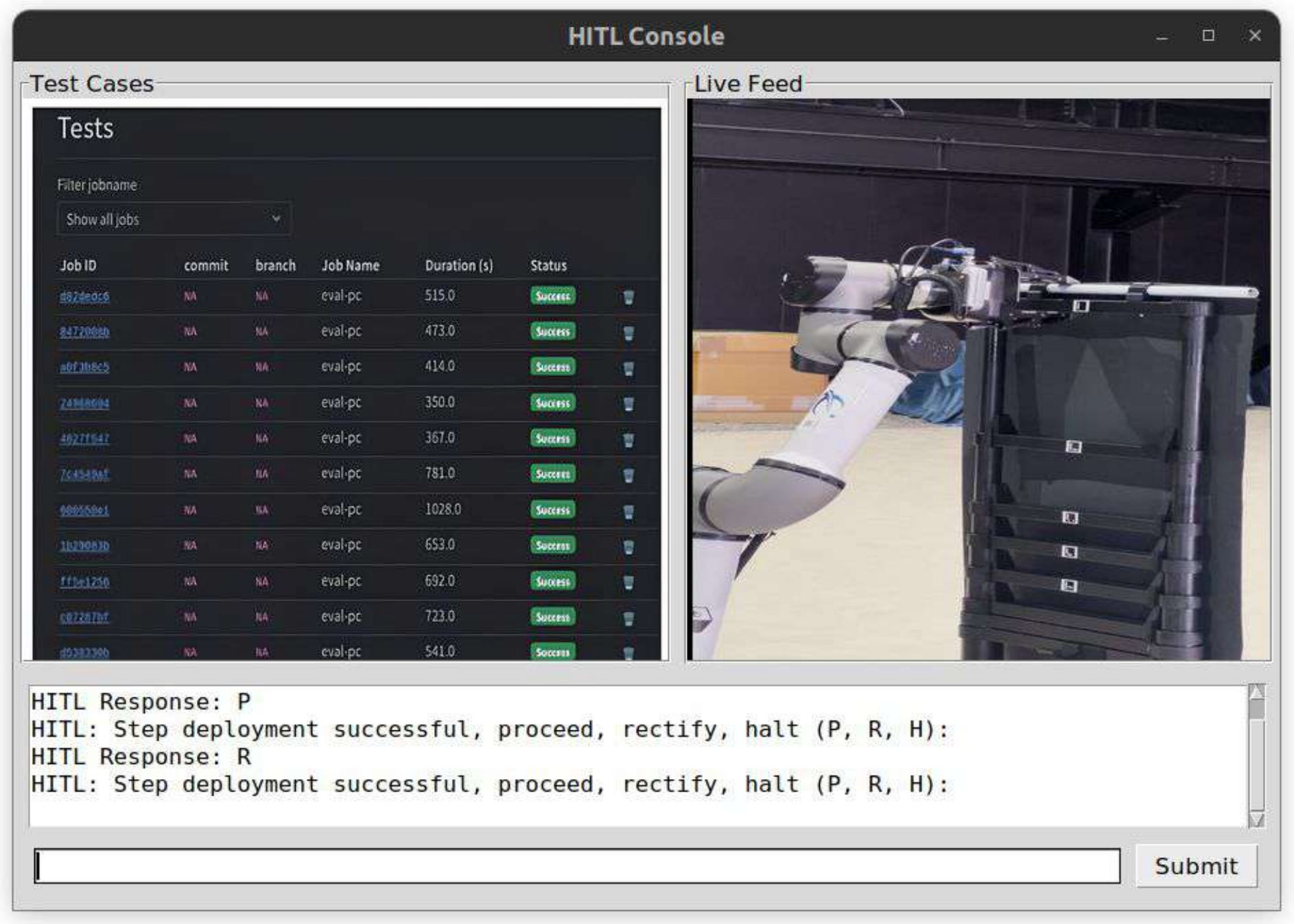}
    \caption{Continuous Integration and Operator GUI}
    \label{fig:GUI}
 \end{figure}
 
\begin{figure*}[t]
    \centering
    \includegraphics[width=0.9\linewidth]{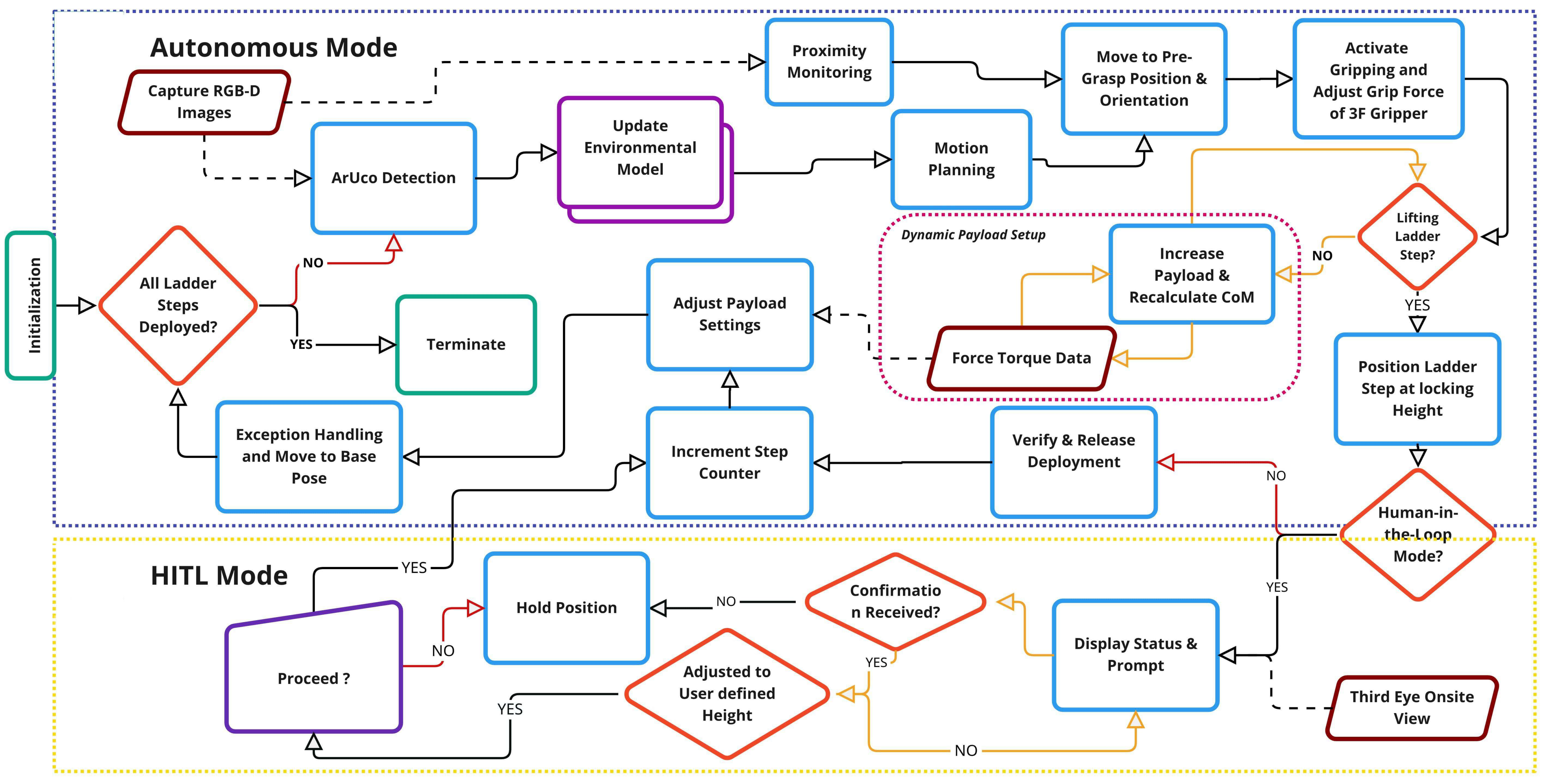}
    \caption{Experiment Procedure Flow Chart}
    \label{fig:exp_procedure}
\end{figure*}

The control system integrates vision, force-torque sensors, and the digital twin for coordinated manipulation. It was verified using the Continuous Integration and Deployment (CI/CD) framework within the ARTEFACTS platform \cite{artefacts, louis_fabian}, which provides configurable standard tests for navigation and manipulation. These cloud-based tests run in parallel, eliminating local execution constraints and enabling extensive simulations. With potentially hundreds of hours of daily simulation, this approach ensures efficient, scalable, and comprehensive system evaluation across diverse scenarios. The results are logged and displayed in a dashboard as shown in Fig.~\ref{fig:GUI} and prompt the operator input to confirm safe operations. 
The hardware enables smooth transitions between autonomous operation and HITL intervention, ensuring flexibility and reliability in complex deployments.

\subsection{Experiment Procedure}\label{experiment_procedure}

Experiments were conducted to validate the autonomous deployment of the extendable ladder-like flexible solar panel, integrating real-time feedback for precise manipulation. Vision-guided pose estimation begins by detecting the ArUco markers on the ladder steps to compute position $(x,y,z)$ and tilt angle for localization. Based on this data, the motion planner executes the inverse kinematics to compute joint configurations while optimizing energy-efficient trajectories using RRT-based path planning. The grasping and lifting phase involves aligning the robotic arm with the step’s tilt, after which the gripper approaches and securely grasps the step, dynamically adjusting grip force based on the force-torque sensor feedback. Once the step is grasped, the deployment phase executes a controlled vertical lift, continuously updating payload parameters and recalculating the center of mass to ensure stability. Real-time monitoring through sensory feedback ensures correct step placement, with HITL intervention correcting discrepancies. This iterative approach minimized positional deviations and trajectory errors, ensuring precise ladder deployment, as depicted in Fig. \ref{fig:exp_procedure}.

\begin{figure}[!htpb]
    \centering
    \includegraphics[width=0.9\columnwidth]{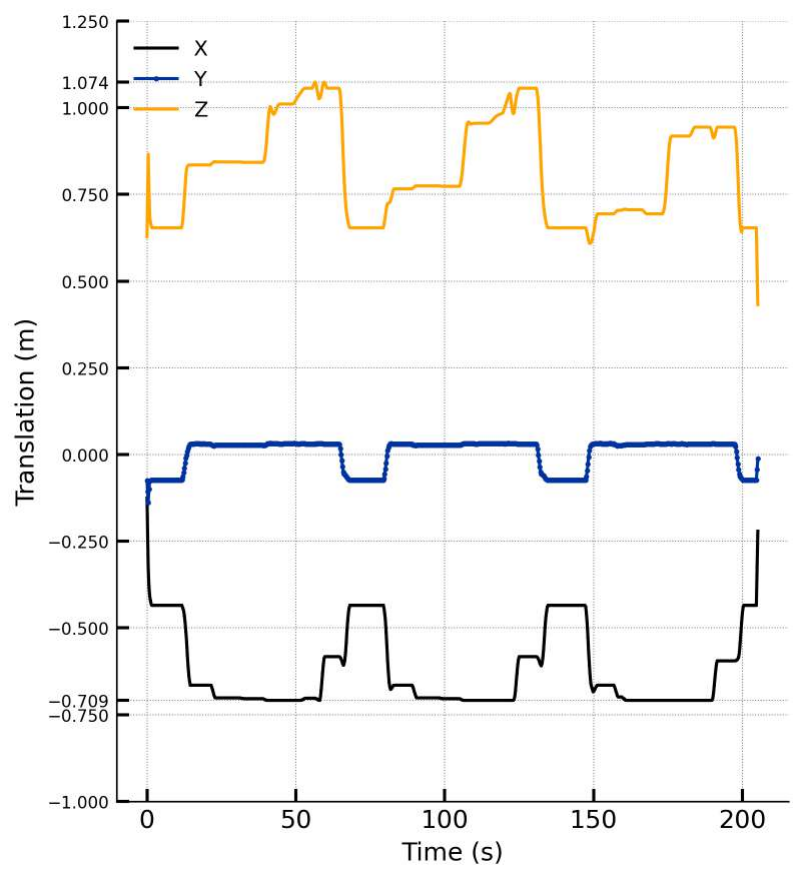}
    \caption{Results of in-lab testing of Autonomous Deployment }
    \label{lab_results}
\end{figure}

\begin{figure*}[t]
    \centering
    \includegraphics[width=\linewidth]{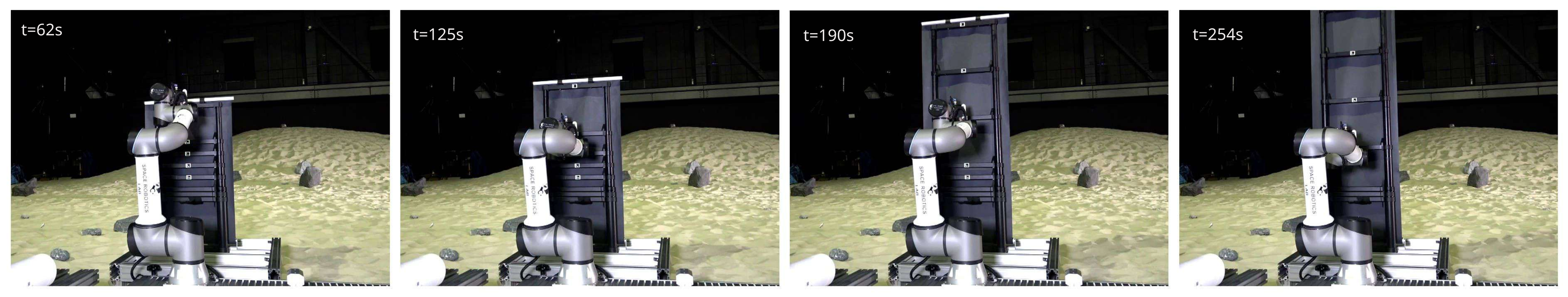}
    \caption{Step-wise deployment of extendable solar panel on the field}
    \label{fig:field_experiments}
\end{figure*}

\begin{figure*}[!htpb]
    \centering
    \includegraphics[width=\textwidth]{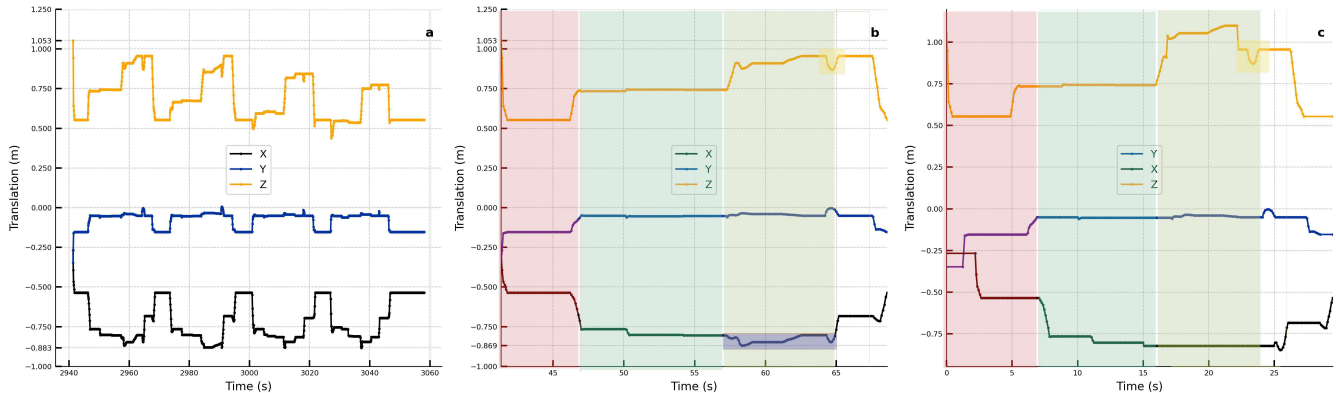}
    \caption{End Effector Trajectory: (a) full deployment over time; (b) Zoomed view of the first step; (c) Action correction using Sequential Axis Motion Planning (SMP) to mitigate deviations due to sand sinkage.}
    \label{fig:combined}
\end{figure*}

\section{Results and Analysis}\label{results} 

\subsection{Validation and in-lab Testing}

The autonomous deployment system was first tested in the lab environment and simulations prior to field tests at JAXA facility. Initial tests were conducted in a controlled $\SI{3}{m}\times\SI{4}{m}$ sandpit of \SI{0.08}{m} depth to simulate lunar surface. Data collection was managed using ROS2 logging joint angles, force-torque data, RGB-D streams, and HITL interventions for post-experiment analysis and system refinement. Simulations and lab tests confirmed reliable performance, with minimal sand sinkage ensuring self-locking operation and stable lighting supporting ArUco detection. The results validated multi-axis motion planning, with consistent self-locking engagement, demonstrating the algorithm’s effectiveness. \textbf{}The graph in Fig.~\ref{lab_results} plots the pose translations in $x$, $y$ and $z$ axes with respect to the manipulator base for the steps deployment. As evident from the graph, the distance of ladder steps remains constant throughout the deployment stage, hence proving vertical deployment resulting in $100\%$ success rate.  These results confirm its readiness for testing in field conditions.

\subsection{Field Testing at JAXA}
\label{subsec:field_testing}

The test field at JAXA provides a cutting-edge environment for simulating extraterrestrial conditions. The facility features a $\SI{20}{m}\times\SI{20}{m}$ test area, filled with over \SI{0.3}{m} of sandy soil closely resembling the granular properties of lunar regolith \cite{Tansa_X}. The field was partially illuminated by a xenon lamp, simulating natural sunlight at lower intensity. This setup assesses robotic systems under lunar-like conditions, including terrain irregularities, reduced traction, and extreme lighting. Figure~\ref{fig:field_experiments} illustrates the step-wise experimental procedure conducted in this facility.

During the field trials, significant deviations in the end-effector trajectory were observed when deploying the ladder on loose, sandy terrain. As illustrated in Fig.~\ref{fig:combined}(a), the complete deployment over time revealed fluctuations in the \(x\)-axis whenever the ladder was raised (i.e., when the tool’s \(z\)-axis position increased). These fluctuations highlight how small vibrations in a single axis can be amplified by simultaneous multi-axis motion, ultimately shifting the base reference of the ladder relative to the manipulator.

Figure~\ref{fig:combined}(b) offers a magnified view of the first step of the deployment. In this figure, the red, cyan, and green regions represent home positioning, alignment and grasping, and the final deployment phase, respectively. Of particular note is the indigo region, which indicates an unintended \(x\)-axis deviation after the end-effector has grasped the ladder. Ideally, once grasping is complete, the \(x\)-axis should remain constant. However, due to diagonal forces introduced by simultaneous axis movements, the ladder tilted in the sand, resulting in sinkage and compromising the self-locking mechanism.

\subsubsection{Mitigating Sand Sinkage through Single-Axis Sequential Motion Planning}
\label{subsec:sequential_axis}

In order to address these deviations, a single-axis \emph{Sequential Motion Planning (SMP)} strategy was implemented at a reduced operational speed of \SI{2}{cm/s}. Rather than moving multiple axes at once, the manipulator actuates each axis sequentially, as shown in Fig.~\ref{fig:combined}(c). Specifically, the robot first aligns along the \(z\)-axis using ArUco markers, then adjusts the \(y\)-axis for precise end-effector positioning, and finally refines the \(x\)-axis alignment before closing the gripper. Only after these steps are verified does the robot lift along the \(z\)-axis to complete the ladder deployment.

As can be seen in Fig.~\ref{fig:combined}(c), this single-axis approach drastically reduces the diagonal forces that cause sand sinkage, thereby mitigating the undesired \(x\)-direction drift. The small bulge marked in yellow denotes a minor Human-in-the-Loop (HITL) correction, which showcases the system’s ability to make fine real-time adjustments. By limiting motion to one axis at a time and operating at lower speeds, the SMP approach preserves the ladder’s self-locking feature, thereby enhancing reliability and safety in off-Earth solar tower deployments.

\section{Discussion}\label{discussions}
Lunar surface operations present multiple uncertainties such as sand sinkage and difficult illumination conditions. This research shows that trajectory deviations during manipulation can be mitigated using sequential axis motion planning and HITL intervention, enhancing stability and reliability. Vision and force-torque sensor feedback improve accuracy, enabling real-time adjustments to grip forces and payload dynamics, while allowing operators to address misalignments and ensure proper attachments on unstable surfaces. However, limitations remain, including the absence of reduced gravity, extreme thermal fluctuations, and vacuum conditions in testing, along with the mechanical constraints of the UR16e robotic arm and potential sensor degradation due to dust. Although lab and field tests demonstrated the system’s robustness in lunar-like conditions, key factors such as lunar dust, reduced gravity, extreme temperatures, and vacuum environments may significantly influence mission performance. While the tests offered valuable insights into deployment precision and sinkage mitigation, future research should incorporate comprehensive environmental simulations and long-term operational assessments to ensure system reliability under actual lunar conditions.

\section{Conclusions}\label{conclusion}

\begin {comment}
This research delineates a robust methodological framework for enhancing robotic manipulation within the context of space applications, particularly under conditions of environmental uncertainty and complexity. The integration of Human-in-the-Loop control mechanisms with autonomous robotic functionalities has proven to be an effective paradigm for navigating the unpredictable challenges of extraterrestrial deployment scenarios. The successful deployment of flexible solar panels via an extendable ladder-like structure underscores the efficacy of precise motion planning, sensory feedback integration, and real-time human oversight.

The implementation of sequential axis motion planning has been pivotal in mitigating the deleterious effects of sand sinkage and trajectory deviations, thereby ensuring deployment stability and accuracy. The utilization of ArUco markers and force-torque sensors furnished reliable real-time data streams that facilitated dynamic error detection and corrective interventions. The system’s performance, as validated through rigorous field testing at JAXA, substantiates its viability for future lunar missions.

This study acknowledges the limitations inherent in the current experimental setup, particularly concerning environmental simulations and mechanical constraints. Future research endeavors will aim to surmount these limitations by incorporating more comprehensive environmental simulations, including lunar gravity and thermal conditions, and by enhancing the mechanical resilience and adaptability of the robotic platform. Additionally, longitudinal operational data will be collected to evaluate the system’s performance over extended mission durations. In summation, the hybrid approach delineated in this study—integrating autonomous robotic operations with HITL control—constitutes a promising trajectory for the reliable and precise assembly of space infrastructure. The insights garnered from this research contribute significantly to the field of space robotics, laying a foundational framework for subsequent innovations in autonomous extraterrestrial construction technologies.
\end{comment}

This study presents a structured approach in autonomous robotic manipulation for extraterrestrial applications, with improved precision, robustness, and adaptability in uncertain space environments. By integrating robotic hardware, control software, motion planning, vision-based localization, limited-sensory feedback, and HITL supervision, this work advances precise autonomous deployment of space infrastructure. The single-axis sequential motion planning technique mitigates challenges such as sand sinkage and trajectory deviations, ensuring structural stability during deployment. Sensory feedback from vision and force-torque sensors enables real-time error detection and correction, enhancing reliability in dynamic conditions. The field tests confirmed the system’s adaptability under lunar-like conditions, with HITL interventions resolving operational ambiguities. However, limitations in environmental simulation and mechanical constraints necessitate future advancements, including lunar gravity modeling, tactile sensing, and enhanced hardware resilience. The hybrid HITL approach lays the groundwork for scalable robotic systems that can build lunar and planetary infrastructure.

\section*{Acknowledgement}
The authors would like to thank Asteria ARTEFACTS for their assistance in setting up the simulation environments.

\bibliographystyle{IEEEtran}
\bibliography{reference.bib}

\end{document}